\documentclass{llncs}
\usepackage[utf8]{inputenc}

\usepackage{graphicx}
\usepackage[utf8]{inputenc}
\usepackage[numbers]{natbib}
\bibpunct{[}{]}{,}{n}{}{;}
\usepackage{amsmath}
\usepackage{amssymb}
\usepackage{geometry}
\usepackage{authblk}
\begin{document}

\title{\textbf{Can Machines Design? An Artificial General Intelligence Approach}}
\author{Andreas M. Hein\inst{1} \and
Hélène Condat\inst{2}}

\institute{Laboratoire Genie Industriel, CentraleSupélec, Université Paris-Saclay,
Gif-sur-Yvette, France
\email{andreas-makoto.hein@centralesupelec.fr}
\and Initiative for Interstellar Studies (i4is), Charfield, UK}

\maketitle

\begin{abstract}
Can machines design? Can they come up with creative solutions to problems and build tools and artifacts across a wide range of domains? Recent advances in the field of computational creativity and formal Artificial General Intelligence (AGI) provide frameworks for machines with the general ability to design. In this paper we propose to integrate a formal computational creativity framework into the Gödel machine framework. We call the resulting framework design Gödel machine. Such a machine could solve a variety of design problems by generating novel concepts. In addition, it could change the way these concepts are generated by modifying itself. The design Gödel machine is able to improve its initial design program, once it has proven that a modification would increase its return on the utility function. Finally, we sketch out a specific version of the design Gödel machine which specifically addresses the design of complex software and hardware systems. Future work aims at the development of a more formal version of the design Gödel machine and a proof of concept implementation.

\end{abstract}

\section{Introduction}
Can machines design? In other words, can they come up with creative solutions to problems \citep{Simon1974} and intervene into their environment by, for example, building tools and artifacts, or better versions of themselves \citep{Myhill1964, Fallenstein2014}? Surprisingly, this question has not received a lot of attention in the current debate on artificial intelligence, such as in Bostrom \citep{Bostrom2014} and Russell \citep{Russell2015}. An exception is the literature in formal artificial general intelligence (AGI) research \citep{Orseau2012,Orseau2011,Soares2014,Fallenstein2014}. If artificial intelligence is going to have a large impact on the real world, it needs to have at least some capacity to create "new" things and to change its environment. The capacity to create new things has also been called "generativity" in the design theory literature \citep{Hatchuel2011}. Such machines could be used across many contexts where the ability to design in the widest sense is required, for example, designing industrial goods such as the chassis of a car that can subsequently be manufactured. Another application could be in space colonization where local resources are used for building an infrastructure autonomously for a human crew \citep{Hein2016f,Hein2014d,Hein2012b}. 
Traditionally, the wider question of creative machines has been treated in the computational creativity community. The computational creativity community has come up with numerous systems that exhibit creativity \citep{Elgammal, Todd1992, Colton, Cope2005, Collaborative}, i.e. systems that are able to conceive artifacts that are considered as novel and creative by humans and/or are novel compared to the underlying knowledge base of the system. Wiggins \citep{Wiggins} and Cherti \citep{Cherti2018} have explored the link between artificial intelligence and creativity. More specifically, Wiggins \citep{Wiggins} formalizes the notions of exploratory creativity and transformational creativity from Boden \citep{Boden} in an artificial intelligence context. A creative system that exhibits exploratory creativity is capable of exploring a set of concepts according to a set of rules. Transformational creativity is by contrast exhibited by a system that can modify the set of concepts itself and / or the rules according to which it searches for the set of concepts. 

At the same time, the artificial general intelligence community is working on general foundations of intelligence and providing frameworks for formally capturing essential elements of intelligence. Within this community, intelligence is primarily defined as general problem-solving \citep{Hutter2004, Intelligence}. According to Goertzel, \citep{Intelligence}, the field of Artificial General Intelligence deals with "the creation and study of synthetic intelligences with sufficiently broad (e.g. human-level) scope and strong generalization capability..." A relevant research stream in this field is the development of the "universalist approach" that deals with formal models of general intelligence. Examples are Hutter's AIXI \citep{Hutter2004}, Schmidhuber's Gödel Machine \citep{Schmidhuber2009}, and Orseau and Ring's space-time embedded intelligence \citep{Orseau2012}. These formal models are based on reinforcement learning where an agent interacts with an environment and is capable of self-improvement. 

In this paper we attempt to integrate Wiggins' formal creativity framework \citep{Wiggins} into an Artificial General Intelligence (AGI) framework, the Gödel machine \citep{Schmidhuber2009}. The purpose is to demonstrate that the mechanisms of self-improvement in AGI frameworks can be applied to a general system design problem. The resulting design Gödel machine designs according to certain rules but is capable of changing these rules, which corresponds to exploratory and transformational creative systems in Wiggins \citep{Wiggins}. Based on this generic framework, we will sketch out a machine that can design complex hardware or software systems. Such systems encompass most products with a high economic value such as in aerospace, automotive, transportation engineering, robotics, and artificial intelligence. 

\section{Literature Survey}

In the literature survey, we will focus on the literature on design theory, formal modeling languages in systems and software engineering, computational creativity, and artificial general intelligence.

The design theory literature provides criteria for how to evaluate a design theory. Hatchuel et al. \citep{Hatchuel2011} introduce two criteria: generativity and robustness. Whereas generativity is the capacity of a design theory to explain or replicate how new things are created, robustness is understood as how sensitive the performance of the designs is with respect to different environments. The main contribution of the design theory literature to a general designing machine are the different forms of generativity and criteria for evaluating design theories.

One possibility to capture generativity is by using a formal design language. Formal design languages belong to the formalized subset of all design languages that are used for generating designs. Formal languages consist of a set of symbols, called alphabet \(\Sigma\), a set of rules, called grammar, that define which expressions based on the alphabet are valid, and a mapping to a domain from which meaning for the expressions is derived \citep{Harel}. This mapping is called “semantics”. The set of all words over \(\Sigma\) is denoted \(\Sigma^*\). The language \(L\) is a subset of \(\Sigma^*\) and contains all expressions that are valid with respect to a grammar. 

For example, programming languages consist of a set of expressions such as 'if' conditions and for-loops. These expressions are used for composing a computer program. However, the expressions need to be used in a precise way. Otherwise the code cannot be executed correctly, i.e. the program has to be grammatically correct. 

According to Broy et al. \citep{Broy2010a}, formal semantics can be represented in terms of a calculus, another formalism (denotational and translational semantics), and a model interpreter (operational semantics). Existing formal semantics for complex systems and software engineering seem to be based on denotational semantics where the semantic domain to which the syntax is mapped is based on set theory, predicate logic \citep{Broy1992, Broy2010}, algebras \citep{Herrmann}, coalgebras \citep{Golden2013} etc.

Formal design languages are formal languages that are used for designing, e.g. for creating new objects or problem-solving. For example, programming languages are used for programs that can be executed on a computer.

The computational creativity literature presents different forms of creativity and creativity mechanisms \citep{Gero1996}. It distinguishes between several forms of creativity, which have been introduced by Boden \citep{Boden}: Combinational creativity is creativity that is based on the combination of preexisting knowledge. For example, the game of tangram consists of primitive geometric shapes that are combined to form new shapes. Exploratory creativity is “the process of searching an area of conceptual space governed by certain rules” \citep{Riedl2006}. Finally, transformational creativity “is the process of transforming the rules and thus identifying a new sub-space.” \citep{Riedl2006} These categories seem to correspond with the generativity categories combinatorial generation, search in topological proximity, and knowledge expansion in design theory \citep{Hatchuel2011}. All three forms of creativity can be generated by computational systems today \citep{Boden}. However, a key limitation is that these systems exhibit these forms of creativity only for a very narrow domain such as art, jokes, poetry, etc. No generally creative system exists. 

The artificial general intelligence literature does seldom treat creativity explicitly. Schmidhuber \citep{Schmidhuber2006,Schmidhuber2010,Schmidhuber2012} is rather an exception. He establishes the link between a utility function and creativity. A creative agent receives a reward for being creative. Hutter \citep{Hutter2004} briefly mentions creativity. Here, creativity is rather a corollary of general intelligence. In other words, if a system exhibits general intelligence, then it is necessarily creative. In the following, we will briefly introduce the Gödel machine AGI framework that has received considerable attention within the community.

\section{Creativity and the Gödel Machine: A Design Gödel Machine}

We use the computational creativity framework from Wiggins \citep{Wiggins} and integrate it with the Gödel machine framework of a self-referential learning system.  In his influential paper, Wiggins \citep{Wiggins} introduces formal representations for creative systems that have been informally introduced by Boden \citep{Boden}, notably exploratory and transformational creativity. We choose the Gödel machine as our AGI framework, as its ability to self-modify is a key characteristic for a general designing machine. Furthermore, it uses a formal language, which makes it easier to combine with formal design languages. However, we acknowledge that AIXI \citep{Hutter2004} and Orseau and Ring's space-time embedded intelligence \citep{Orseau2012} should be considered for a similar exercise. 

A Gödel machine that can generate novel concepts (paintings, poems, cars, spacecraft) is called design Gödel machine in the following. Such a machine is a form of creative system, defined as a "collection of processes, natural or automatic, which are capable of achieving or simulating behaviour which in humans would be deemed creative." \citep{Wiggins}

\begin{figure}[htbp]
\centering
\includegraphics[scale=0.2]{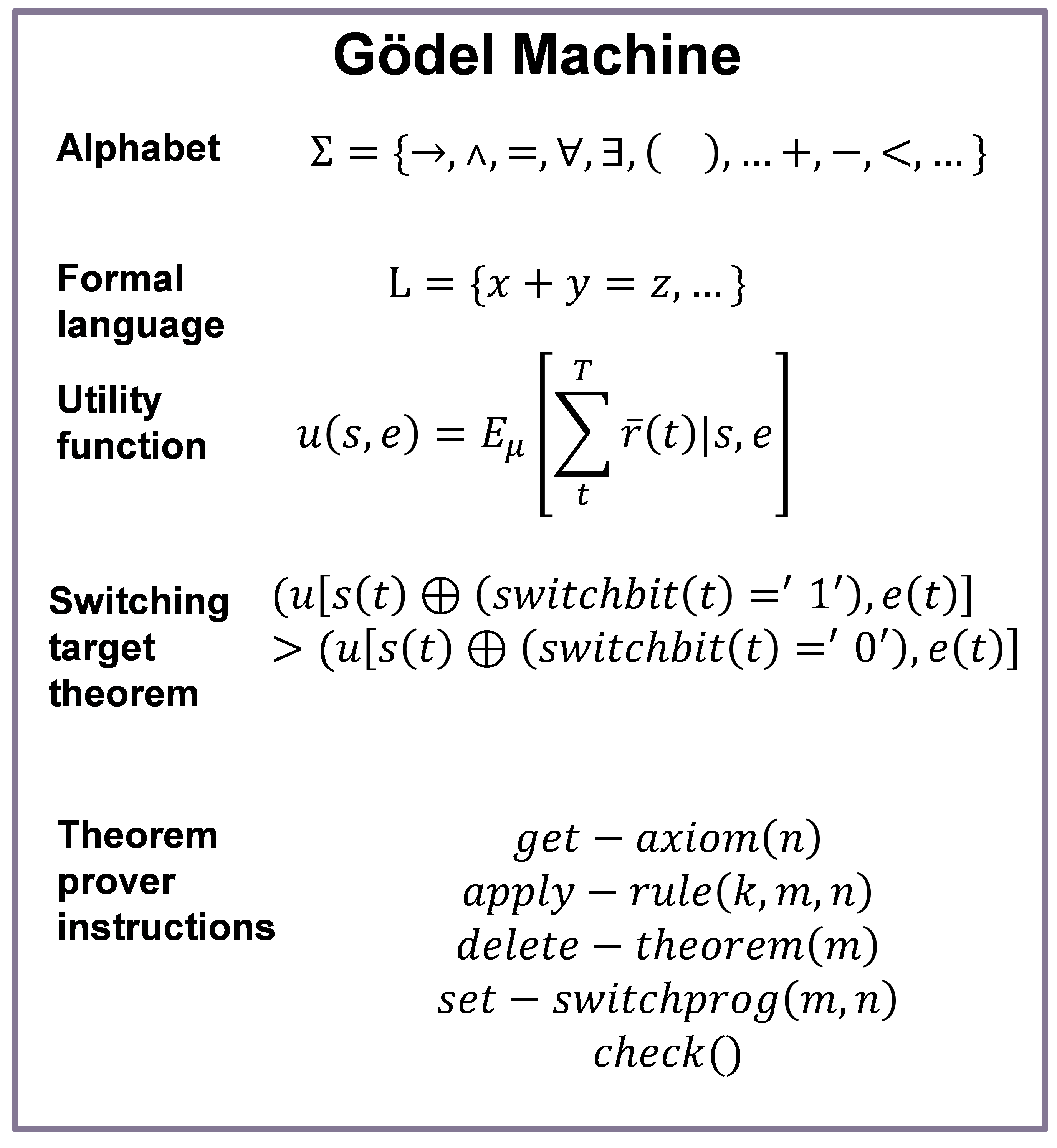}
\caption{Elements of the Gödel machine}
\label{fig:3}
\end{figure}

The original Gödel machine consists of a formal language \(\mathcal{L}\) that may include first order logic, arithmetics, and probability theory, as shown in Fig. \ref{fig:3}. 

It also consists of a utility function \(u\) whose value the machine tries to maximize. 

\[u(s,e):\mathcal{S}\times \mathcal{E}\rightarrow \mathbb{R} \quad \]

\begin{equation}
u(s,e)=E_\mu [\sum_{\tau=time}^T r( \tau ) |s,e] \quad for \quad 1\leq t \leq T
\end{equation}

Where \(s\) is a variable state of the machine, \(e\) the variable environmental state, \(r(t)\) is a real-valued reward input at a time \(t\). \(E_\mu (\cdot|\cdot)\) denotes the conditional expectation operator of a distribution \(\mu\) of a set of distributions \(M\), where \(M\) reflects the knowledge about the (probabilistic) reactions of the environment. 

How does the Gödel machine self-improve? A theorem prover searches for a proof that a modification can improve the machine’s performance with respect to the utility function. Once a proof is found that a modified version of itself would satisfy the target theorem in Equation (\ref{eq:1}), the program \(switchprog\) rewrites the machine's code from its current to its modified version. The target theorem essentially states that when the current state \(s\) at \(t_1\) with modifications yields a higher utility than the current machine, the machine will schedule its modification.

\begin{equation} \label{eq:1}
(u[s(t_1 ) \oplus switchbit(t_1 ) ='1'), Env(t_1 )] > u[s(t_1) \oplus switchbit(t_1) ='0'), Env(t_1)])
\end{equation}

The basic idea of combining the Gödel machine framework with the formal creativity framework of Wiggins \citep{Wiggins} is to construct a Gödel machine where its problem solver corresponds to an exploratory creative system and the proof searcher corresponds to a transformational creative system. The transformational creative system can modify the exploratory creative system or itself. 

More formally, the design Gödel machine consists of an initial software \(p(1)\). \(p(1)\) is divided into an exploratory creative system which includes an initial policy \(\pi(1)_{env}\), which interacts with the environment and a transformational creative system, which includes an initial policy \(\pi(1)_{proof}\). \(\pi(1)_{proof}\) searches for proofs and forms pairs of \((switchprog, proof)\), where the proof is a proof of a target theorem that states that an immediate rewrite of \(p\) via \(switchprog\) would yield a higher utility \(u\) than the current version of \(p\). \(\pi(1)_{env}\) is more specifically interpreted as a set of  design sequences comprising design actions. A design sequence, for example, is the order in which components are combined to form a system. The different ways of how components can be combined are the design actions and the sequence of how they are combined is the design sequence. 

The design Gödel machine consists of a variable state \(s \in \mathcal{S} \). The variable state \(s\) represents the current state of the design Gödel machine, including a set of concepts \(c(t)\) at time \(t\) that the machine has generated, a set of syntactic and knowledge-based rules \(\mathcal{R}\) that define the permissible concepts in a design language \(\mathcal{L}\), and a set of sequences of design actions \(\pi_{env}\) for generating concepts and getting feedback for these concepts from the environment. The machine generates concepts in each time step \(t\), including the empty concept \(\top\). It receives feedback on the utility of these concepts via the utility function \(u(s,e):\mathcal{S}\times \mathcal{E} \to \mathbb{R} \), which computes a reward from the environmental state \(e \in \mathcal{E}\). Analogous to the exploratory creative system in Wiggins \citep{Wiggins}, \(\pi_{env}\) and \(u\) are part of a 7-tuple \(<\mathcal{U}, \mathcal{L}, [.], \langle.,.,.\rangle, \mathcal{R}, \pi_{env}, u >\), where \(\mathcal{U}\) is a universe of concepts, \([.]\) is an interpretation function that applies the syntactic and knowledge-based rules \(\mathcal{R}\) to \(\mathcal{U}\), resulting in the set of permissible concepts \(\mathcal{C}\). The interpreter \(\langle.,.,.\rangle\) takes a set of concepts \(c_{in}\) and transforms them into a set of concepts \(c_{out}\) by applying \(\langle \mathcal{R},\pi_{env}, u \rangle \):

\begin{equation}
(c_{out})= \langle \mathcal{R}, \pi_{env},u \rangle (c_{in})
\end{equation}

Self-modification for the design Gödel machine means that parts of the exploratory creative system and transformational creative system can be modified. Regarding the former, the transformational creative system is able to modify the exploratory creative system's rules \(\mathcal{R}\), the sequences of design actions \(\pi_{env}\), and the utility function \(u\). For this purpose, the transformational creative system searches for a proof that a modification would lead to a higher value on the meta utility function \(u_{meta}\). By default, \(u_{meta}\) returns 0 if the target theorem in Equation (\ref{eq:1}) is not satisfied and 1 if it is. If the target theorem is satisfied, this modification is implemented in the subsequent time step. 
In addition, a target theorem \(u_{meta}\) could capture criteria for a good design sequence in \(\pi_{env}\) that are expected to lead to a higher value on \(u\). Examples are measures for the originality of the created designs via a design sequence, if originality is expected to lead to higher values on \(u\). The proof searcher \(\pi_{proof}\) that searches for the proof and the proof itself are expressed in a meta-language \(\mathcal{L}_{meta}\). The proof is based on axioms, rules, and theorems in \(\mathcal{R}\) and \(\pi_{env}\), the meta-language syntax and rules \(\mathcal{R}_{meta}\), and the proof strategies \(\pi_{proof}\) of the proof searcher. Hence, the transformational creative system can be expressed as the 7-tuple:

\begin{equation}
<\mathcal{L}, \mathcal{L}_{meta}, [.]_{meta}, \langle .,.,. \rangle_{meta}, \mathcal{R}_{meta}, \pi_{proof}, u_{meta}>
\end{equation}

More specifically, in case \(u\) is not modified, the proof searcher \(\pi_{proof}\) generates pairs of \(\mathcal{R}\) and \(\pi_{env}\) from an existing \(\mathcal{R}\) and \(\pi_{env}\) by applying an interpreter \(\langle .,.,. \rangle_{meta}\) with \(\mathcal{R}_{meta}\), \(\pi_{proof}\), and \(u_{meta}\): 

\begin{equation}
(\mathcal{R}_2, \pi_{env2})= \langle \mathcal{R}_{meta}, \pi_{proof},u_{meta} \rangle_{meta} (\mathcal{R}_1, \pi_{env1})
\end{equation}

This formulation is similar to the transformational creative system in Wiggins \citep{Wiggins}. If the proof searcher can prove \(u_{meta}((\mathcal{R}_2, \pi_{env2}),e_1) > u_{meta}((\mathcal{R}_1, \pi_{env1}),e_1)\), the design Gödel machine will switch to the new rules \(\mathcal{R}_2\) and design sequences \(\pi_{env2}\). 

Analogous to the original Gödel machine, the transformational creative system in the design Gödel machine is capable of performing self-modifications, for example, on the proof searcher and the meta-utility function: 

\begin{equation}
(\mathcal{X}_2)= \langle \mathcal{R}_{meta}, \pi_{proof},u_{meta} \rangle_{meta} (\mathcal{X}_1)
\end{equation}

where \(\mathcal{X}\) is one of the elements in \(<\mathcal{L}, \mathcal{L}_{meta}, [.], \langle .,.,. \rangle, \mathcal{R}_{meta}, \pi_{proof}, u_{meta}>\). Self-reference in general can cause problems, however, as Schmidhuber \citep{Schmidhuber2009} notes, in most practical applications, they are likely not relevant. A design Gödel machine would start with an initial configuration and then modify itself to find versions of itself that yield higher values on its utility function.

\begin{figure}[htbp]
\centering
\includegraphics[scale=0.2]{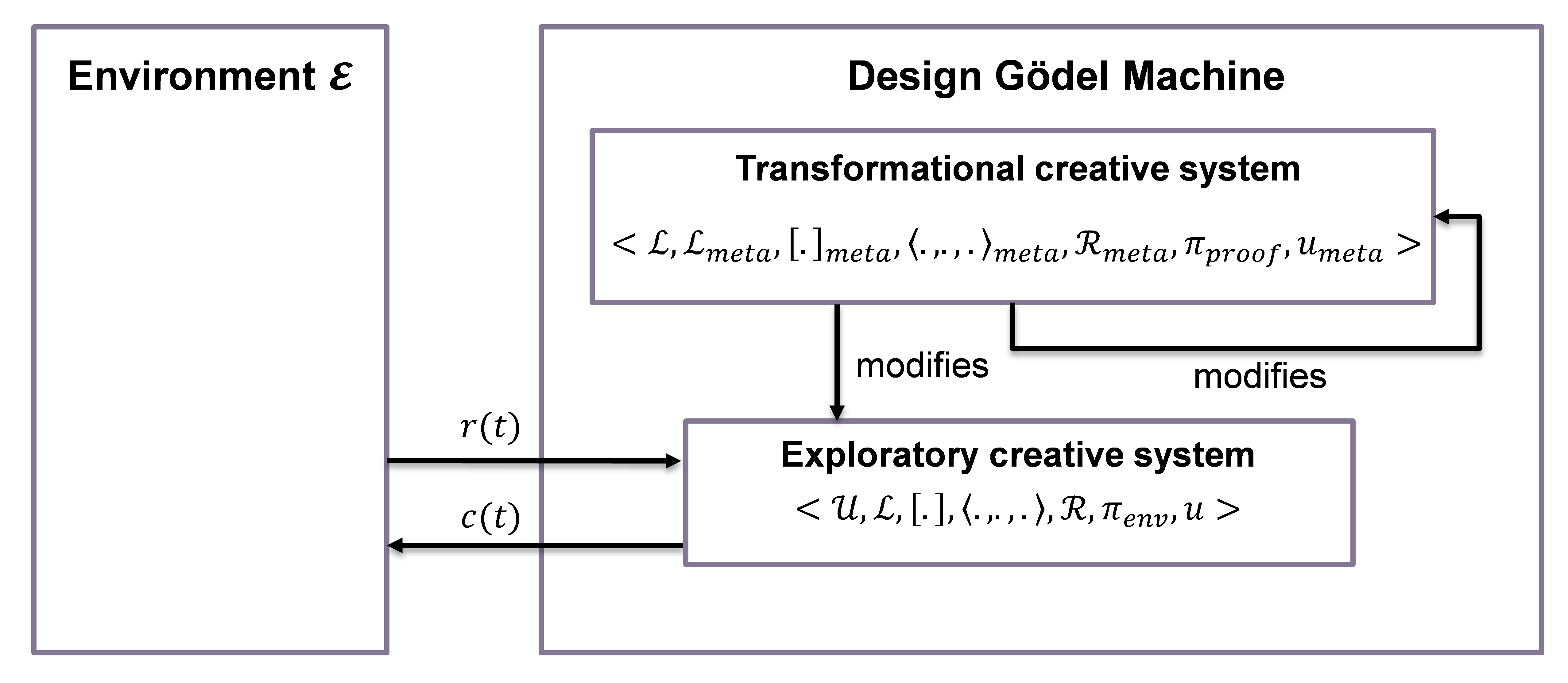}
\caption{Elements of the design Gödel machine}
\label{fig:4}
\end{figure}

Fig. \ref{fig:4} provides an overview of the main elements of the design Gödel machine that have been introduced before.

\section{A Design Gödel Machine for Complex Systems Design}

A specific version of the design Gödel machine for designing complex software and hardware systems can be imagined. It would include a set of syntactic and knowledge-based rules \(\mathcal{R}\) that define sound designs (concepts for hardware and software) in the specific domain and a set of design actions such as abstraction, refinement, composition, and verification \citep{Broy1992, Broy2010, Golden2013} that can be combined into design sequences \(\pi_{env}\). The environment \(\mathcal{E}\) could be a virtual test environment or an environment in which design prototypes are tested in the real world.  

Important principles of formal systems and software engineering are components and their interactions, abstraction, composition, refinement, and verification \citep{Broy1992, Broy2010, Golden2013}. Broy \citep{Broy1992, Broy2010} defines interactions in terms of streams and interfaces. Golden \citep{Golden2013} defines interactions in terms of dataflows. Component functions are specified in terms of transfer functions that transform inputs into outputs. The component behavior is specified in terms of state machines. Golden \citep{Golden2013} specifies component behavior via a timed Mealy machine, Broy \citep{Broy1992, Broy2010}  uses a state-oriented functional specification for this purpose. 

Apart from this basic representation of a system as a set of interacting components, abstraction, composition, refinement, and verification are important principles during the design of a system. 

Abstraction means that details are left out in order to facilitate the comprehension of a complex system, reduce computational complexity or for mathematical reasons \citep{Golden2013}. Abstraction is treated by Golden \citep{Golden2013} via dataflow, transfer function, and component abstraction. He remarks that abstraction can also lead to non-determinism due to the underspecification of the abstracted system. 

Composition is the aggregation of lower-level components together with their interactions to higher-level components. Herrmann et al. \citep{Herrmann} propose a compositional algebra for aggregating components. Broy \citep{Broy2010} specifies composition as the assignment of truth values to system-level inputs and outputs based on component-level inputs and outputs. Golden \citep{Golden2013} divides composition into product and feedback. His notion of product is similar to the compositional algebra in Herrmann et al. \citep{Herrmann} and defines products of dataflows, transfer functions, and components. Feedback further deals with outputs of a component that are fed into the same component as an input. 

Refinement is the addition of details to arrive from a general to a more specific system specification. Golden \citep{Golden2013} defines refinement as a form of decomposition, which is the inverse operation of composition. Broy \citep{Broy2010} defines different forms of refinement: property, glass box, and interaction refinement. Both Golden and Broy interpret refinement as an addition of properties and decomposition of components / interactions.

Verification is the process of checking requirements satisfaction. Golden \citep{Golden2013} assigns requirements to a system or component via "boxes" that specify the system or component's inputs, outputs, and behavior. Broy \citep{Broy1992, Broy2010} similarly distinguishes between global (system-level) requirements and local (component-level) requirements. The verification process in his case is essentially formally proving that the system and its components satisfy the requirements. 

The literature on formal modeling languages for software and systems engineering provides the necessary semantics and rules for describing complex soft- and hardware systems. However, the main shortcoming of formal modeling languages for complex software and hardware systems is that they cannot generate these systems by themselves. In other words they are not generative without additional generativity mechanisms and a knowledge base.

\subsection{Design Axioms}

As in the original Gödel machine, theorem proving requires a enumerable set of axioms. These axioms are strings over a finite alphabet \(\Sigma\) that includes symbols from set theory, predicate logic, arithmetics, etc. The design Gödel machine for complex systems design includes a number of design-related axioms that will be presented in the following. 
The design axioms belong to three broad categories. 
The first are axioms related to the formal modeling language, describing its abstract syntax (machine-readable syntax), the semantic domain, for example, expressed in predicate logic, and the semantic mapping between the abstract syntax and the semantic domain. The semantic domain and mapping in Golden \citep{Golden2013} and Broy \citep{Broy1992, Broy2010}  can be essentially reformulated in terms of set theory, predicate logic, arithmetics, and algebra. These axioms belong to \(\mathcal{R}\), but specifically define which designs are "formally correct". We denote the set of these axioms as \(\mathcal{R}_{formal}\). These axioms include formal definitions for a system, component, interfaces, and interactions between component etc. 

The second category consists of axioms related to different mechanisms of generating designs. Specifically, these are axioms for refinement, abstraction, composition, verification, and axioms that describe domain-specific rules based on domain-specific knowledge. We consider these axioms as part of the set of design sequences \(\pi_{env}\).

The third category are axioms that describe conceptual knowledge such as the notion of "automobile". Without being too restrictive, such conceptual knowledge would include axioms for parts and whole, i.e. mereological statemenents \citep{Simons1987}. For example, an automobile has a motor and wheels. The axioms also belong to \(\mathcal{R}\), however, contrary to \(\mathcal{R}_{formal}\), they are not general principles of representing complex systems but knowledge specific to certain concepts. Such axioms are expressed by \(\mathcal{R}_{ck}\). 

\subsection{System}

According to Golden \citep{Golden2013}, a system is a 7-tuple \(\int=<\mathbb{T}_s,Input,Output,S,q_0,\mathcal{F},\mathcal{Q}>\) where \(\mathbb{T}_s\) is a time scale called the time scale of the system, \(Input=(In,\mathcal{I})\) and \(Output=(Out,\mathcal{O})\) are datasets, called input and  output datasets, \(S\) is a nonempty \(\epsilon\)-alphabet, called the \(\epsilon\)-alphabet of states, \(q_0\) is an element of \(S\), called the initial state, \(\mathcal{F}:In\times S \times \mathbb{T}_s \rightarrow Out\) is a function called functional behavior, \(\mathcal{Q}:In\times S \times \mathbb{T}_s \rightarrow S\) is a function called states behavior. \((Input, Output)\) are called the signature of \(\int\). This definition of a system corresponds to a timed Mealy machine \citep{Mealy1955}.

It is rather straight-forward to model the Gödel machine in this system framework, if the loss in generality of using the timed Mealy machine is considered acceptable. In that case, we take: \(\mathbb{T}_s=\mathbb{N}\), \(Input=(\mathcal{E},\mathbb{E})\), \(q_0 = s(t_1)\), \(Output=(\mathcal{S},\mathbb{S})\), \(\mathcal{F}:  \mathcal{E} \times \mathcal{S} \times \mathbb{T}_s \rightarrow \mathcal{A}\), \(\mathcal{Q}: \mathcal{E} \times \mathcal{S} \times \mathbb{T}_s \rightarrow \mathcal{S}\). \(\mathbb{E}\) and \(\mathbb{S}\) are any data behaviors on \(\mathcal{E}\) and \(\mathcal{S}\) respectively. 

Formulating the design Gödel machine in the system framework allows for applying the formal machinery of the framework such as refinement, abstraction, verification etc. that the design Gödel machine can apply to itself.

\subsection{Refinement and Abstraction}
Refinement and abstraction relate system representations that are at different levels of abstraction \citep{Broy1992, Golden2013}. According to Broy \citep{Broy2010}, refinement may include the addition of properties to the system that makes it more restrictive, or includes its decomposition into components. For example:

\begin{equation}
x \implies y \circ z
\end{equation}

where the system \(x\) is decomposed into the components \(y\) and \(z\).

\subsection{Composition}

The composition operator is important for combining components into a system with their respective interfaces. A generic composition operator can be understood as:

\begin{equation}
y \otimes z \implies x 
\end{equation}

where the components \(y\) and \(z\) are composed to \(x\). 
These operators would not only need to be defined for software systems, such as proposed by \citep{Golden2013, Broy1992, Broy2010} but would also need to include interpretations of the composition for physical systems \citep{Herzig2010}. This is likely to entail mereological questions of parts and wholes \citep{Simons1987}. 

\subsection{Verification}

We interpret verification in two distinct ways: First, with respect to a set of requirements \(\Phi\) that is part of the environment \(\mathcal{E}\), where \(\mathcal{E}\) returns a reward input \(r(t)\) to the design Gödel machine. Based on \(r(t)\) and the respective set of concepts \(\mathcal{C}\), the utility function \(u\) is evaluated. Such a utility function would have the form \(\tilde{u}: \mathcal{C} \times \Phi \rightarrow \mathbb{R}\), with \(\mathcal{C} \subset \mathcal{S}\) and \(\Phi \subseteq \mathcal{E}\). 

Second, the set of requirements \(\Phi\) is internal to the design Gödel machine. The requirements describe expectations with respect to the environment \(\mathcal{E}\). Specifically, the satisfaction of the requirements is expected to return a reward input \(r(t)\) from the environment. For example, if a concept \(c\) (a car) exhibits a property \(a\) (consumes less than 3 l/km in fuel), then the resulting \(r(t)\) will result in a higher \(u\) than for a different property (consumes 10 l/km of fuel). The conditional expectation operator \(E_\mu (\cdot|\cdot)\) from the original Gödel machine is slightly modified for this purpose, leading to a utility function \(u: \mathcal{C} \times \Phi \times \mathcal{E} \rightarrow \mathbb{R}\). 

\begin{equation}
u(c,\varphi,e)=E_\mu [\sum_{\tau=time}^T r( \tau ) | c,\varphi,e] \quad for \quad 1\leq t \leq T
\end{equation}

where \(\varphi \in \Phi\) and \(c \in \mathcal{C}\). The requirements \(\Phi\) are themselves expectations of what the environment \(\mathcal{E}\) "wants" from the design(s). They are subject to modifications, depending on the environment's response \(r(t)\). This second interpretation of verification captures nicely the distinction between verification and validation in systems engineering, where verification checks if the design satisfies the requirements and validation checks if the requirements were the right ones \citep{Haskins2007}.

\section{Limitations}

Design Gödel machines are subject to the same limitations as the original Gödel machine \citep{Schmidhuber2009} such as the Gödel incompleteness theorem \citep{Godel1931} and Rice's theorem \citep{Rice}. 

Apart from these theoretical limitations, a basic limitation of the design Gödel machine presented here is that it is based on a formal language. Computing systems that are not based on a formal language could not be addressed by this approach. 

As Orseau \citep{Muehlhauser2013} has remarked, the Gödel machine is expected to be computationally extremely expensive for reasonably complex practical applications. 

An important limitation of this paper is that we have not provided an implementation of the design Gödel machine together with a proof of concept demonstration. This remains a task for future work. Furthermore, for an application in a real-world context, the problem to be solved by the machine needs to be carefully selected. For example, which tasks based on which inputs and outputs are interesting for automation \citep{Rigger2018}? Apart from the possibility of proper formalization, economic criteria will certainly play an important role. 

\section{Conclusions}

In this paper, we proposed to integrate a formal creativity framework from Wiggins into the Gödel machine framework of a self-referential general problem solver. Such an integration would be a step towards creating a "general designing machine", i.e. a machine that is capable of solving a broad range of design problems. We call this version of the Gödel machine a design Gödel machine. The design Gödel machine is able to improve its initial design program, once it has proven that a modification would yield a higher utility. The main contribution of this paper to the artificial general intelligence literature is the integration of a framework from computational creativity into an artificial general intelligence framework. In particular, exploratory and transformational creative systems are integrated into the Gödel machine framework, where the initial design program is part of the exploratory creative system and the proof searcher is part of the transformational creative system. Of particular practical interest would be a design Gödel machine that can solve complex software and hardware design problems. Elements of such a machine are sketched out. However, a practical implementation would require a more extended formal systems engineering framework than those existing today. An interesting area for future work would be the integration of Wiggin's framework into other artificial general intelligence frameworks such as Hutter's AIXI and Orseau and Ring's space-time embedded intelligence.

\bibliography{main}
\end{document}